\documentclass{article}
\usepackage[utf8]{inputenc}

\title{Procedural Kernel Networks}
\author{Bartlomiej Wronski}
\date{Google Research}

\usepackage[square,sort,comma,numbers]{natbib}
\usepackage{lipsum}
\usepackage{amsmath, xparse}

\usepackage{booktabs}

\usepackage{graphicx}
\usepackage{caption}
\usepackage{hyperref}
\usepackage{multicol}
\usepackage{float}
\usepackage[a4paper, total={6.8in, 9.82in}]{geometry}
\newenvironment{Figure}
  {\par\medskip\noindent\minipage{\linewidth}}
  {\endminipage\par\medskip}

\begin{document}

\hyphenpenalty=1000
\captionsetup[figure]{labelfont={bf},name={Figure},labelsep=period}

\maketitle
\vspace{0.1cm}
\begin{figure*}[ht]
  \vspace{-0.2in}
  \centering
  \includegraphics[width=0.7\linewidth]{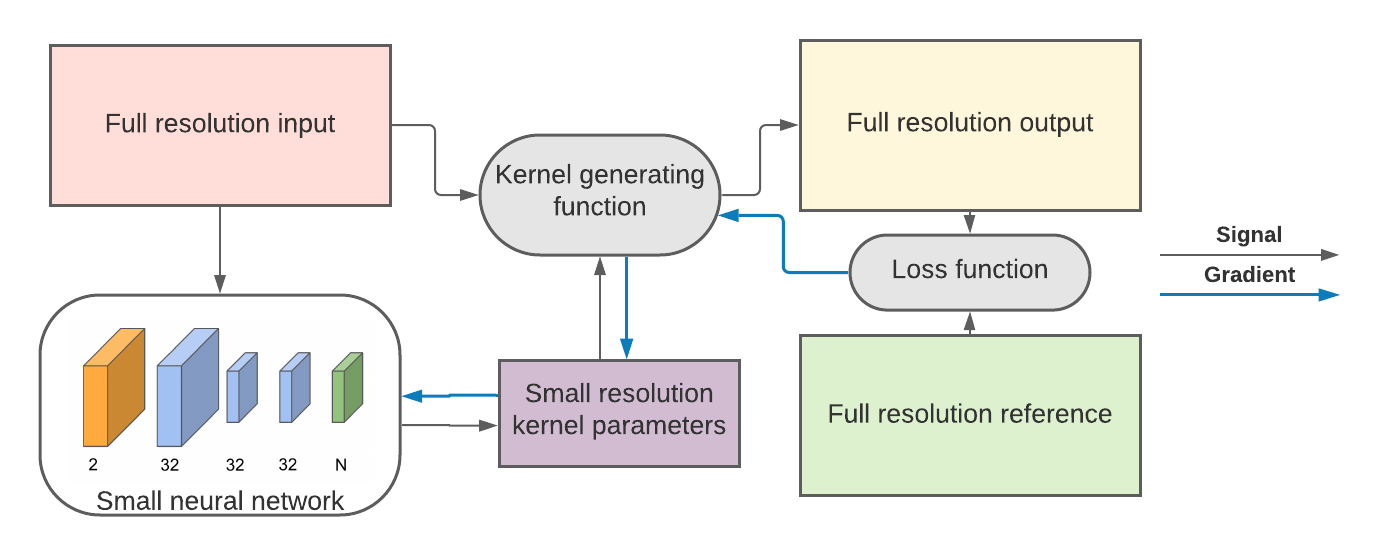}
  \caption{\textbf{Overview of our method.}
  In contrast to image predicting or kernel prediction networks, our method uses small (shallow, aggressively spatially pooling aggregated information) neural networks to predict a low resolution image describing local kernel generating parameters. Those parameters are used to compute final linear or non-linear kernels that are applied to the input image to produce the final output. Our method relies on differentiable kernel generator.}
  \label{fig:overview}
  \vspace{-0.1in}
\end{figure*}

\begin{abstract}
In the last decade Convolutional Neural Networks (CNNs) have defined the state of the art for many low level image processing and restoration tasks such as denoising, demosaicking, upscaling, or inpainting.
However, on-device mobile photography is still dominated by traditional image processing techniques, and uses mostly simple machine learning techniques or limits the neural network processing to producing low resolution masks.
High computational and memory requirements of CNNs, limited processing power and thermal constraints of mobile devices, combined with large output image resolutions (typically 8--12 MPix) prevent their wider application.
In this work, we introduce Procedural Kernel Networks (PKNs), a family of machine learning models which generate parameters of image filter kernels or other traditional algorithms.
A lightweight CNN processes the input image at a lower resolution, which yields a significant speedup compared to other kernel-based machine learning methods and allows for new applications.
The architecture is learned end-to-end and is especially well suited for a wide range of low-level image processing tasks, where it improves the performance of many traditional algorithms.
We also describe how this framework unifies some previous work applying machine learning for common image restoration tasks.
\end{abstract}

\begin{multicols}{2}
\section{Introduction}
In the last decade, CNNs demonstrated great performance in many image processing and restoration tasks like denoising, demosaicking, and inpainting.
While these techniques are attractive, smartphone photography is largely dominated by traditional image processing techniques based on signal processing concepts, manually designed heuristics, and tuning by experts.
Most commercial vendors do not disclose details of their image signal processing (ISP) algorithm details, but for the denoising task existing literature and open source libraries describe the practical use of methods like bilateral filter \cite{hasinoff2016burst}, or wavelet shrinkage techniques \cite{dcraw}.
The computational and memory requirements of CNNs for large output image resolutions (typically 8--12 MPix) are incompatible with the limited processing power and thermal constraints of mobile device.
Even the authors of popular commercial packages targetting desktop platforms offer the CNN based techniques only as an experimental, slow processing option \cite{adobe}.
Machine learning is typically used in a shallow and limited form, for example by learning direct filters \cite{romano2016raisr}.
Most successful approaches using deep learning on mobile devices either predict auxiliary images at a much lower resolution \cite{wadhwa2018synthetic}, or they generate coefficients for a low resolution bilateral grid that approximates other image processing operations \cite{chen2007real}.

Kernel Predicting Networks (KPNs) \cite{bako2017kernel} are a type of CNN, originally developed for denoising Monte Carlo ray traced images in movie production.
KPNs allowed for smaller network architectures, less visual artifacts, and much faster training time due to the implicit regularization provided by kernel prediction \cite{vogels2018denoising}.

We expand on the ideas of kernel prediction networks and the deep bilateral grid, and propose to connect and generalize those in a framework we call \textbf{Procedural Kernel Networks (PKNs)}.
We use networks to predict \textit{functional} kernels with a number of parameters much smaller than the number of weights of the equivalent discrete filter.
PKNs don't rely on fixed kernel sample positions and can be predicted at any resolution.
Taken to the extreme, kernel parameters could be predicted globally per image.

We show how a network order of magnitude smaller than contemporary deep learning approaches provides significant quality improvements over hand-tuned, traditional image processing algorithms.


In this work, we propose and analyze a general, learned, procedural kernel framework that:
\begin{itemize}
\item Improves the quality of domain specific, highly optimized and specialized traditional algorithms or hardware.
\item Decouples network processing resolution from the final image resolution---ranging from full image resolution to a single predicted value.
\item Decouples the effective kernel spatial support from the parameter count.
\item Produces interpretable kernel parameter maps that can be analyzed or further tuned by human experts.
\item Is able to produce continuous kernels, not relying on fixed pixel or point locations for operation on sparse inputs, or continuous resampling.
\end{itemize}

\section{Background}

\subsection{Kernel predicting networks}
Convolutional Neural Networks have been used successfully for common image processing operations like denoising, demosaicking, upsampling, or inpainting and have been shown to reach or exceed performance of traditional algorithms on the benchmark datasets. On the other hand, CNNs predicting pixel colors directly are growing in size. These can require massive amounts of data, with total training time expressed in years, and yet can still produce visual artifacts or demonstrate low performance when exposed to test data outside of the training sets.

Kernel Predicting Networks \cite{bako2017kernel} are an area of active research that addresses many of those shortcomings.
Kernel Predicting networks were shown theoretically and practically to train much faster \cite{vogels2018denoising} than classic pixel predicting networks, they allow for smaller network sizes, and they have been applied to processing bursts of images \cite{mildenhall2018burst}.
Kernel prediction regularizes the output to be a convex combination of the input pixels, guaranteeing no hallucinated colors or brightness levels.
The ideas of kernel prediction has also been applied to the convolution filters inside the network structure itself \cite{jia2016dynamic,kokkinos2019pixel}.

Simultaneously, a different line of work builds on ideas of per-pixel learned filter banks \cite{getreuer2018blade} and explores the use of linear combination of either global, predefined kernel bases for tasks of single image upsampling \cite{li2020lapar}, motion blur deconvolution \cite{carbajal2021single}, or learning per-pixel kernel selection for general image processing \cite{kokkinos2019pixel}.

\subsection{Fast Neural Network processing on mobile devices}
Kernel Predicting Networks help with training convolutional neural networks, their generalization, and producing robust results when presented with novel data, but they don't address high computational cost as compared to traditional algorithms.
This poses a challenge for performance-constrained environments, such as mobile phones, where practical applications of machine learning and CNNs are constrained to either predicting images at a lower resolution \cite{wadhwa2018synthetic}, or single learned filters chosen by handcrafted heuristics \cite{romano2016raisr}.

Gharbi et al. \cite{gharbi2017deep} propose to address this by using a CNN to predict coefficients for a low-resolution bilateral grid, which is then efficiently sampled at the full resolution. This way, inference resolution can be decoupled from the final image resolution and used to approximate different image processing operations.

This idea of predicting smaller resolution coefficients was further expanded to a more general use of a learned guided filter \cite{wu2018fast}, and in recent years to meta-learning and image-to-image translation tasks \cite{shaham2020spatially}.

\begin{figure*}[ht!]
  \centering
  \includegraphics[width=0.8\linewidth]{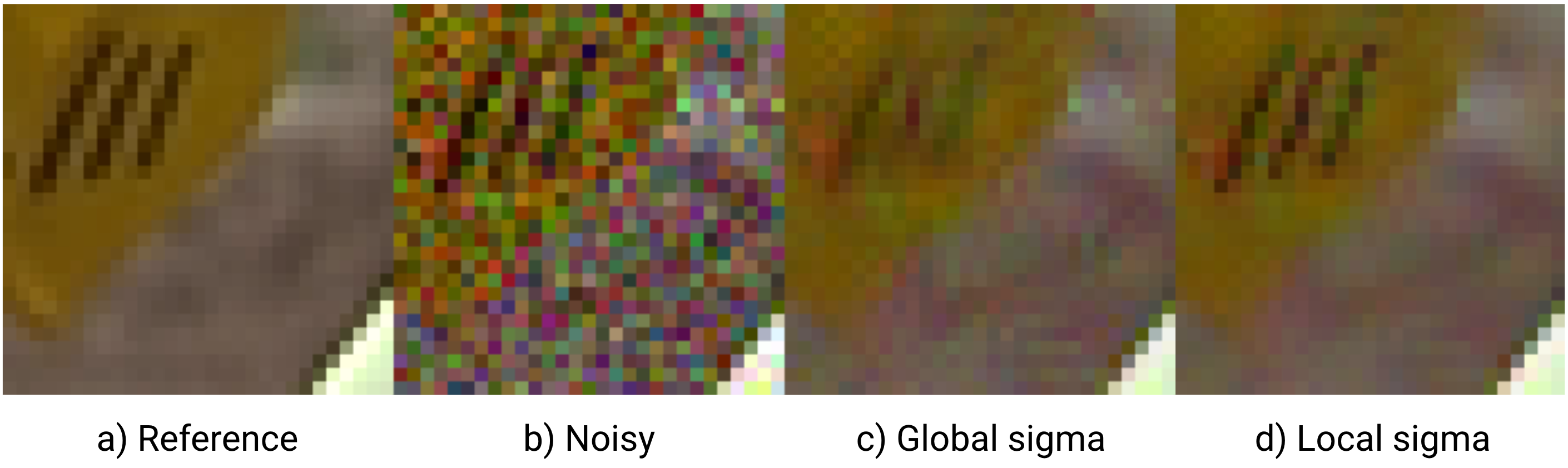}
  \caption{\textbf{Oracle experiment.} A reference image \textbf{a)} corrupted with AWGN \textbf{b)}.
  When denoised with a 5x5 NLM filter with a globally optimal single signal sigma parameter \textbf{c)}, we observe detail loss and leftover noise. Locally adapting sigma \textbf{d)} allows for better detail preservation and noise removal.}
  \label{fig:global_local_bilateral}
\end{figure*}

\subsection{Traditional imaging algorithms in CNN setting}
While most of the literature considers CNNs as a standalone building block for image processing operations, there is some interest in using machine learning to aid traditional algorithms, or vice versa.
One of the earliest examples involves using a single multilayer perception (MLP) for predicting optimal weights for a joint bilateral filter for denoising Monte Carlo images \cite{kalantari2015machine}.
Similarly, convolutional neural networks can be used for using a bilateral grid not only for approximating other image operators \cite{chen2007real}, but also traditional bilateral denoising \cite{meng2020real}.
Recent work proposed to use bilateral filtering with network-predicted features and network-predicted filter parameters for achieving real-time performance on the Monte-Carlo rendering denoising task \cite{isik2021anf}.

One of the advantages of traditional algorithms apart from their computational efficiency can be explainability, or relationship to physical processes.
For this reason, neural networks were successfully used in the field of medical and scientific imaging as priors aiding traditional optimization techniques \cite{diamond2017unrolled}.
Other traditional image and signal filtering operations used successfully inside a neural network setting include Wiener deconvolution \cite{dong2020deep} and guided filtering \cite{wu2018fast}.

A separate line of work uses neural networks to approximate traditional algorithms, including their tunable and parametric behavior \cite{fan2019general}.
Recent work \cite{tseng2019hyperparameter} combines the two ideas, and proposes to build a neural network approximating a black box algorithm or whole image processing unit that can be used in an optimization setting for finding best parameters for different objectives.

%


\section{Our method}
\subsection{Motivation: limitations of global parameters}
We begin with the observation that despite significant advancement in image denoising over the last fifteen years (using both traditional techniques \cite{dabov2007image} and deep learning \cite{zhang2017beyond}), variants of the classic non-local means filter (NLM) \cite{buades2005non} and bilateral filtering are still widely used in photographic pipelines.
This is the case especially on mobile devices \cite{hasinoff2016burst} due to their computational efficiency, interpretability, robustness, and lack of requirements for careful pipeline and data modeling.
Such simpler algorithms are typically reused by vendors for many years, and are efficient not only because of algorithmic simplicity, but also get implemented in hardware (camera ISP), or using dedicated hardware processors.

Bilateral filtering based denoisers are used in complex pipelines, involve many heuristics and require careful tuning, which is a time consuming task performed by specialized image quality experts.
This motivated using neural networks \cite{tseng2019hyperparameter} and differentiable programming approaches \cite{li2018differentiable} for finding optimal algorithm or tuning parameters.
Those approaches automate and significantly simplify the tuning process to find optimal algorithm parameters, but still involve trade-offs---a globally optimal tuning minimizing the loss function across whole dataset compromises the performance locally---both between example images in the dataset, as well as within a single image.

For our motivation, consider a particular non-local means filter\footnote{Note that in the text we will use NLM and bilateral kernel interchangeably, as a bilateral kernel is a special case of the NLM kernel with a patch size of 1x1.} (3x3 patch size, 5x5 search window, Gaussian falloff) applied separately to each channel of the image.
We take images from Kodak dataset \cite{kodak}, add Gaussian noise (sigma 0.1), fix the filter spatial sigma, and attempt to find optimal signal sigma parameter minimizing the L2 loss (compared to the ground truth) averaged across the whole image.
In this toy experiment, we observe (\autoref{fig:global_local_bilateral}a-c) that globally optimal parameters can lead to locally suboptimal filtering---both too noisy with leftover artifacts, as well as overblurring image details.
This is caused by limitations of the NLM kernel and pixelwise comparisons, that are unable to go beyond simple pixelwise or patchwise comparisons.

\subsection{Oracle experiment with locally optimal parameters}
We hypothesize that the limited image quality is caused by finding a single per-image algorithm parameter and that different regions of the image can benefit from different parameters.
To verify that, we seek ``locally optimal'' parameters through an \textit{oracle} optimization process.

We take pairs of clean and noisy images, and apply a NLM filter with \textit{spatially-varying sigma} on the noisy images. To find the optimal local signal sigma (computed at half of the image resolution to enforce smoothness), we minimize the L2 loss locally compared to the ground truth clean image.
We rely on the differentiability of the NLM kernel with regard to its parameters, which lets us optimize spatially-varying signal sigma using gradient descent and related optimization techniques.

This local optimization leads much better results, both in terms of detail preservation, as well as less leftover noise and visual artifacts (\autoref{fig:global_local_bilateral}a-d). It significantly improves the PSNR by over 1\,dB (\autoref{table:psnr_denoise}).

While this experiment suggests that letting the NLM parameters vary spatially can lead to significantly better results, the approach is not direclty applicable, since when denoising real-world images we don't have access to their clean versions.
\begin{table*}[t]
\centering
  \begin{tabular}{ *{7}{cccc} }
    \toprule
    \textbf{Noise sigma} &  \textbf{PSNR Optimal} & \textbf{PSNR Network} &  \textbf{PSNR Optimal} \\
     &  \textbf{sigma per image} & \textbf{predicted sigma} &  \textbf{local sigma} \\    
    \midrule
    0.025 & 35.93\,dB & 36.25\,dB & 36.68\,dB\\
    0.05 & 31.84\,dB & 32.28\,dB  & 32.87\,dB\\
    0.1 & 27.90\,dB & 28.49\,dB  & 29.03\,dB\\
    0.2 & 23.72\,dB & 24.64\,dB  & 24.99\,dB\\
    \bottomrule
  \end{tabular}
\caption{PSNR comparison of globally optimal and locally predicted tuning values.}
\label{table:psnr_denoise}
\end{table*}

\subsection{Network prediction of locally optimal NLM parameters}\label{bilat_params}
To turn this idea into practical application, we propose to use a simple, low parameter neural network to predict locally good tuning values (\autoref{fig:local_tuning}).
\begin{Figure}
 \centering
 \vspace{0.1in}
 \includegraphics[width=\linewidth]{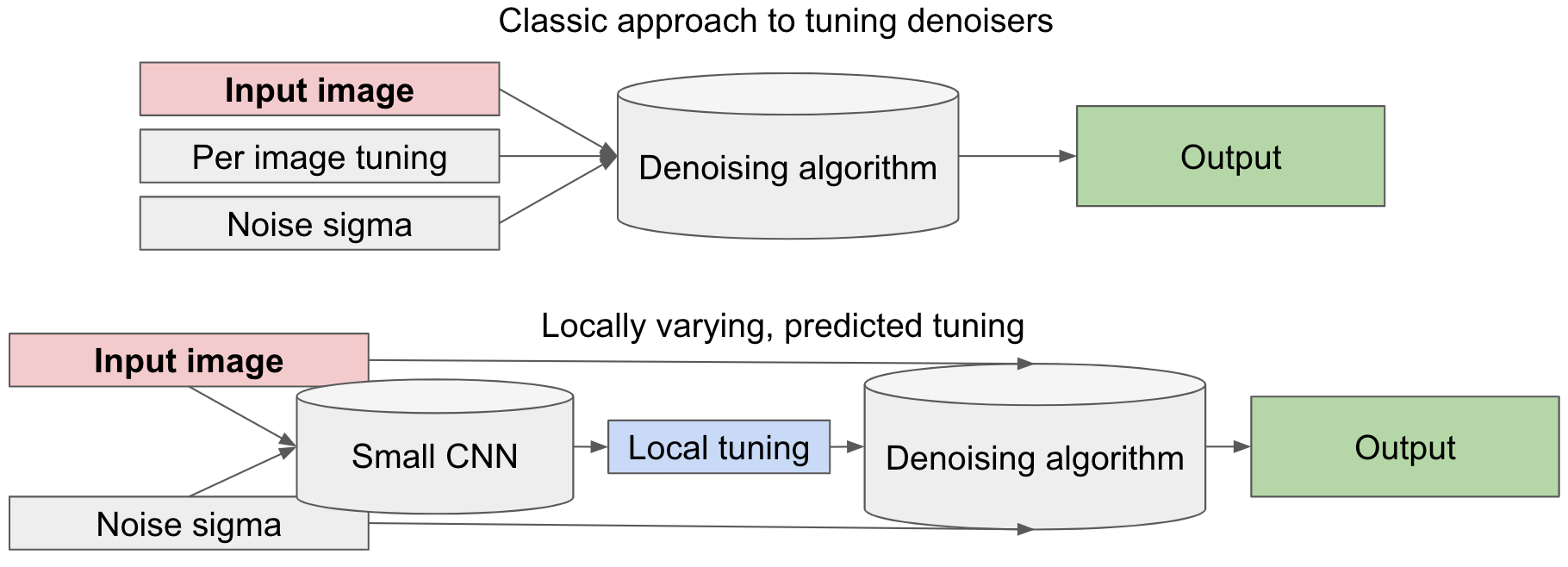}
 \captionof{figure}{
  A small network used to predict \textit{spatially varying} algorithm parameters.
  }
 \label{fig:local_tuning}
 \vspace{0.1in}
\end{Figure}
Note that while some earlier work used networks for predicting optimal channel weights for joint-bilateral filtering based on multi-channel, pixel-wise properties  \cite{kalantari2015machine}, we aim to infer optimal parameters for the traditional bilateral filter based on the neighborhood of the processed pixel.
We train a simple network with the architecture presented in \autoref{fig:simple_net}.

\begin{Figure}
 \centering
 \vspace{0.1in}
 \includegraphics[width=0.55\linewidth]{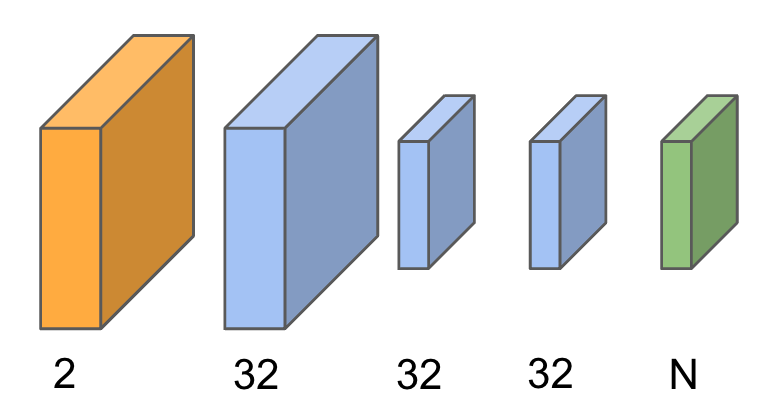}
 \captionof{figure}{
  Network architecture use for NLM parameter prediction. The yellow block is the input (image contents and noise standard deviation), blue blocks represent 3x3 convolution, and the green block outputs N kernel parameters through a 1x1 convolution.
  }
 \label{fig:simple_net}
 \vspace{0.1in}
\end{Figure}

To train the network, used algorithm and its optimized parameters have to be differentiable.
In the case of non-differentiable algorithms, one might use approach described in prior work for creating differentiable algorithm proxies \cite{tseng2019hyperparameter}.

We note that the network itself has less than 20k parameters, and most computations are done in lower resolution than the input image.
Additionally, it has a similar spatial support footprint as the denoiser itself, thus any expected improvements would come purely from learning higher level and structural information.

In this experiment, we feed the network with a single channel of the image to be denoised together with expected per-pixel noise standard deviation.
The network outputs half-resolution signal constrained to range [0,4] using a sigmoid function, which is used as local multiplier on top of the noise standard deviation. This is used as a local spatial sigma for the NLM filter.

We train the network on a subset of 10k random images from Open Images v6 dataset (whole network training until convergence takes less than 10 minutes on a single GPU cloud machine), and use Kodak image dataset as a validation set.
During training, we use random noise standard deviations sampled uniformly from [0.05,0.1]. 

A small network without access to clean images cannot achieve the performance of oracle, but the quantitative improvements on this simple experiment (\autoref{table:psnr_denoise}) are within \textbf{0.32--0.92\,dB} as compared to per-image parameters.
Note that during the training, we never show examples with noise sigma below 0.05 or above 0.1, and yet we observe generalization outside of this range.
We observe also qualitative improvements that we assess as being perceptually in between the locally optimal oracle results, and the per-image optimal tuning parameter (\autoref{fig:oracle_vs_net}).
Given that the oracle experiment is the upper bound of the quality of selected configuration of the NLM filter, we find those results confirming our intuition of local improvements.

\begin{Figure}
 \centering
 \vspace{0.1in}
 \includegraphics[width=\linewidth]{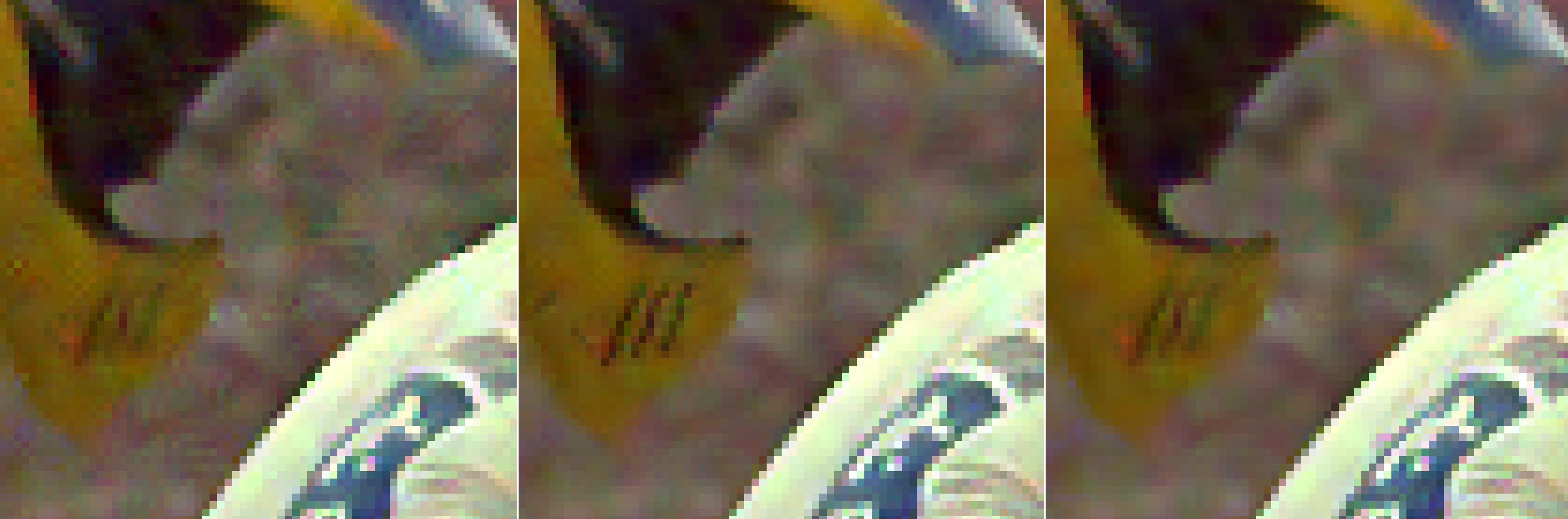}
 \captionof{figure}{
  \textbf{Network parameter prediction.}
  \textbf{Left:} Per-image tuning.
  \textbf{Middle:} Oracle tuning.
  \textbf{Right:} Tuning predicted locally by a small neural network.}
  \label{fig:oracle_vs_net}
  \vspace{0.1in}
\end{Figure}

We analyze the outputs of the network to verify the inferred NLM sigma multiplier maps are spatially smooth and detect salient image features (\autoref{fig:denoise_mask}).
\begin{Figure}
 \centering
 \vspace{0.1in}
 \includegraphics[width=\linewidth]{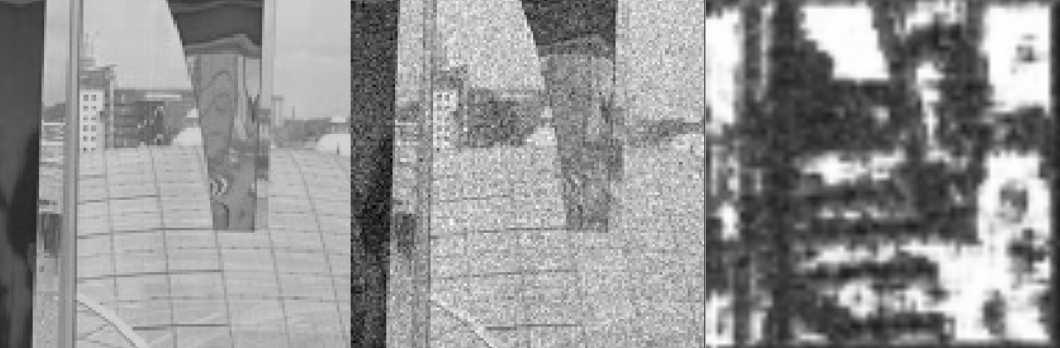}
 \captionof{figure}{
  Demonstration of the produced local parameters.
  \textbf{Left:} Reference image.
  \textbf{Middle:} Noisy image tuning.
  \textbf{Right:} Sigma multiplier parameter image / map.
  Produced parameters delineate smooth areas (more denoising) from edges or textured ones (less denoising).}
  \label{fig:denoise_mask}
  \vspace{0.1in}
\end{Figure}
The inferred parameter maps are interpretable, and can be used to analyze why the denoiser produced more smooth or sharper images in given image and in a given area.
Such produced results can be subject to further tuning by human experts or based on the user preferences without the need for retraining---to adjust the trade-off between denoising and sharpness, one can simply rescale the output parameters, or apply a more general non-linear remapping.
Additionally, the spatial smoothness of the inferred tuning maps suggests that it's possible to reduce inference resolution even further.

In the next section we explain how the idea presented connects prior work on general image filtering and kernel predicting networks.

\subsection{Linear algebra of image filtering}
Given the motivation above, we formalize the results as \textbf{Procedural Kernel Networks} and generalize them to other kernels (beyond NLM and bilateral kernel).

Starting with the bilateral filter kernel:
\begin{equation}
    Y(x) = \frac{1}{\sum W} \sum_{x_i \in \Omega} X(x_i)W_x(x, x_i)W_X(X(x), X(x_i))
\end{equation}
Bilateral filter is a weighted average of pixels around the neighborhood of a destination pixel.
$W_X(x, x_i)$ is a spatial weighting function, and $W_X(X(x), X(x_i))$ is a signal weighting function---both are often Gaussian kernel.
While the weighting functions are non-linear, the bilateral filter is a linear combination of surrounding pixels, where at each pixel we can write down the effective kernel:
\begin{equation}
    Y(x) = \sum_{x_i \in \Omega} X(x_i)w_i 
\end{equation}
\begin{equation}
    w_i = \frac{W_x(x, x_i)W_X(X(x), X(x_i))}{\sum_{x_i \in \Omega} W_x(x, x_i)W_X(X(x), X(x_i))}
\end{equation}
Prior work \cite{milanfar2012tour} has comprehensively explored how most traditional denoising, deconvolution and other image filtering operations can be written in a matrix form $W \cdot X$, where rows and columns of the matrix $W$ correspond to the input and output pixels of the filter. Some filters like the bilateral filter produce sparse $W$ matrix, while for Non-Local Means with an unlimited search window the $W$ matrix is dense in general.

Using this notation, we can treat the bilateral kernel as an output of a function generating matrix entries through spatial and signal relationships of neighboring pixels.

Spatial relationships using a spatially constant function are also used for much simpler kernels, like a box filter, or a Gaussian kernel:
\begin{equation}
    W_x = \frac{1}{\sqrt{2\pi}\sigma}e^{-\frac{\left \| x-x_i \right \|^2}{2\sigma^2}}, W_X=1
\end{equation}

We can express not only standard convolutions, but we can also add a per-element constant term to the input by expanding the dimensionality of the input (mapping to homogeneous coordinates):
\begin{equation}
X' = \begin{bmatrix}
X\\ 
1^T
\end{bmatrix}
\end{equation}
And similarly extending the $W$ matrix by an additional column.

\subsection{Procedural Kernel Networks}
We propose to unify this framework with kernel predicting networks, regressing optimal bilateral filter parameters, and predicting linear combinations of basis functions, and call it \textbf{Procedural Kernel Networks}.

We define the filtering weight of a pixel $x_i$ for the pixel $x$ as a function of spatial pixel relationships, signal levels, and a set of parameters $\Omega$:
\begin{equation}
    W(x, x_i) = F(\Omega(x), x, x_i, X(x), X(x_i))
\end{equation}
Classic \textbf{Kernel Predicting Networks} \cite{bako2017kernel} can be expressed as predicting directly a locally constant weights that depend only on the processed pixel center, and spatial coordinates of the surrounding pixels - identity function mapping:
\begin{equation}
    F = \Omega(x, x_i)
\end{equation}
We contrast the two approaches in \autoref{fig:kpn_vs_pkn}.
\begin{Figure}
 \centering
 \vspace{0.1in}
 \includegraphics[width=\linewidth]{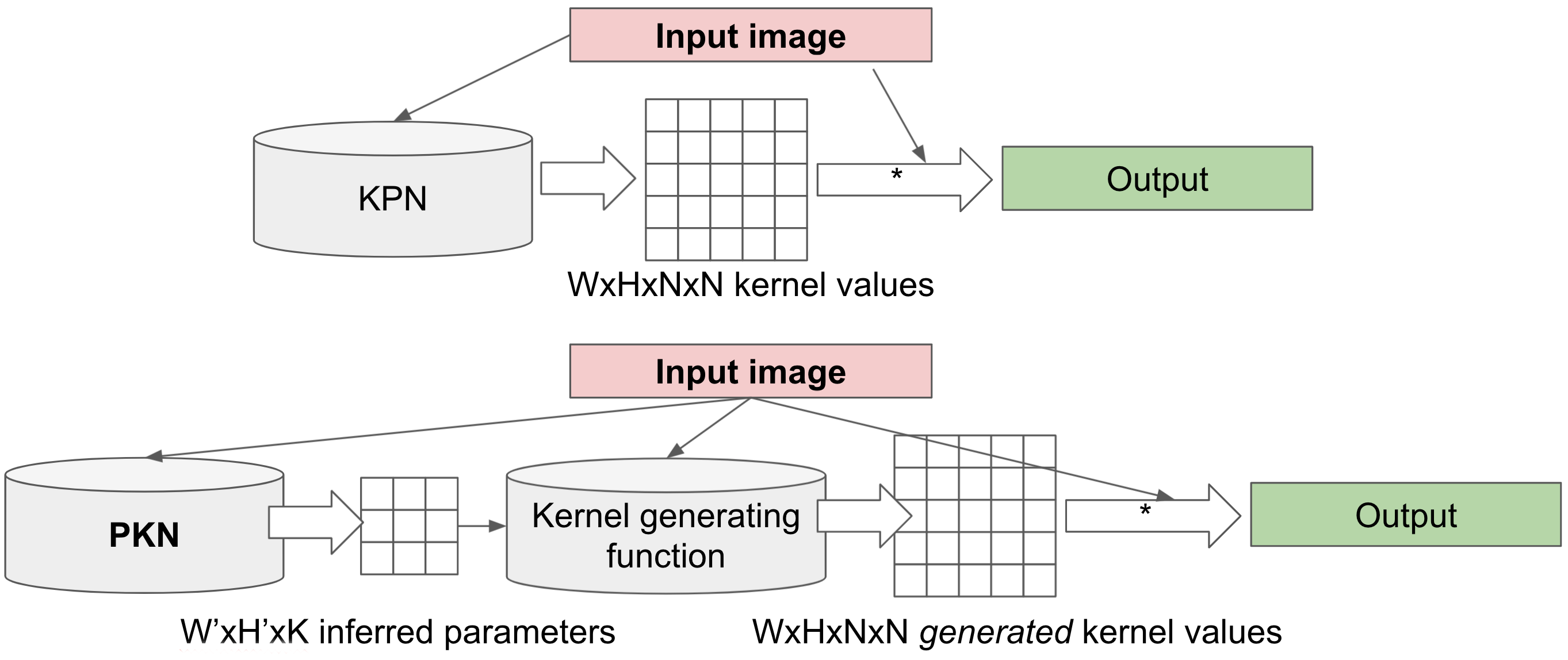}
 \captionof{figure}{KPN (\textbf{top}) as contrasted with PKN (\textbf{bottom}).}
 \label{fig:kpn_vs_pkn}
 \vspace{0.1in}
\end{Figure}

The Procedural Kernel Network framework allows for expressing wide variety of other kernel generating functions and unifies some other prior work.
A network predicted linear combination of basis function \cite{xia2020basis, li2020lapar, carbajal2021single, kokkinos2019pixel} is then defined as: 
\begin{equation}
    F = \sum_n\Omega(x, n) * K_n(x_i)
\end{equation}
Where $K_n$ are different global kernels.

For the experiments proposed in \autoref{bilat_params}, $\sigma$ for the bilateral filter acts as the single $\omega$ parameter of $\Omega$:
\begin{equation}
\omega(x) = \sigma(x)
\end{equation}
Deep bilateral learning \cite{gharbi2017deep} that predicts coefficients of the bilateral grid \cite{chen2007real,isik2021anf} and predicted coefficients of a guided filter \cite{wu2018fast} can be both written:
\begin{equation}
    F = \omega_0(x) + \omega_1(x) X(x)
\end{equation}
Where $\omega_0(x)$ and $\omega_1(x)$ are bi- or trilinearly interpolated.
Those approaches share fast inference, significantly faster training time, and are deployed on mobile platforms.
We observe that this is both because procedural and functional kernels allow for decoupling of the processing resolution from the processed image resolution, as well as provide strong regularization (spatial smoothness and the output space constrained by the kernel parameters).

In the next section we explore using different simple, differentiable kernels for low parameter, learned processing in different applications beyond bilateral denoising.

\section{Different kernels and their applications}
In the following sections, we look at some examples of different procedural kernels and how parameter prediction can be used for a variety of traditional image processing operations.
We use the same network architecture in all of our experiments, only changing the output remapping function and the number of output channels.

\subsection{Anisotropic denoising}\label{aniso_denoise}
The bilateral kernel is not the only kernel weight generating function we can use in the PKN framework for denoising.
Any kind of spatially varying blur kernel could be parameterized to adaptively preserve details or remove noise.

Note that isotroptic Gaussian blur is a degenerate form of the bilateral filter, where the signal sigma is infinity.
Here, we propose using an anisotropic Gaussian blur kernel for edge-preserving behavior.
Such edge-aware kernels are used in applications beyond image restoration, such as 3D mesh reconstruction and fluid particle filtering \cite{yu2013reconstructing}, as well as multi-frame super-resolution \cite{wronski2019handheld}.

An anisotropic Gaussian can be parameterized by the covariance matrix $\Sigma$, or by its inverse, the precision matrix $\Sigma^{-1}$:
\begin{equation}
\Omega(x)=e^{-\frac 1 2 (x_i-x)^\mathrm{T}\Sigma^{-1}(x_i-x)}
\end{equation}
One challenge with parameterizing with full covariance matrices is that directly predicting the matrix entries can lead to a non-invertible matrix.
Similarly, if we tried to optimize precision matrix entries directly, a network could predict an invalid precision matrix.
To address this challenge, we optimize $\sigma_1$, $\sigma_2$, and $\rho$ in the following covariance matrix parameterization:
\begin{equation}
\Sigma=
\begin{bmatrix}
\sigma_1^2 & \rho \sigma_1 \sigma_2\\ 
\rho \sigma_1 \sigma_2 & \sigma_2^2
\end{bmatrix}
\end{equation}
When $\rho$ is constrained to a range [-1, 1] and $\sigma$ values are positive, we are guaranteed to get a correctly formed covariance matrix.
\begin{figure*}[ht]
  \centering
  \includegraphics[width=0.95\linewidth]{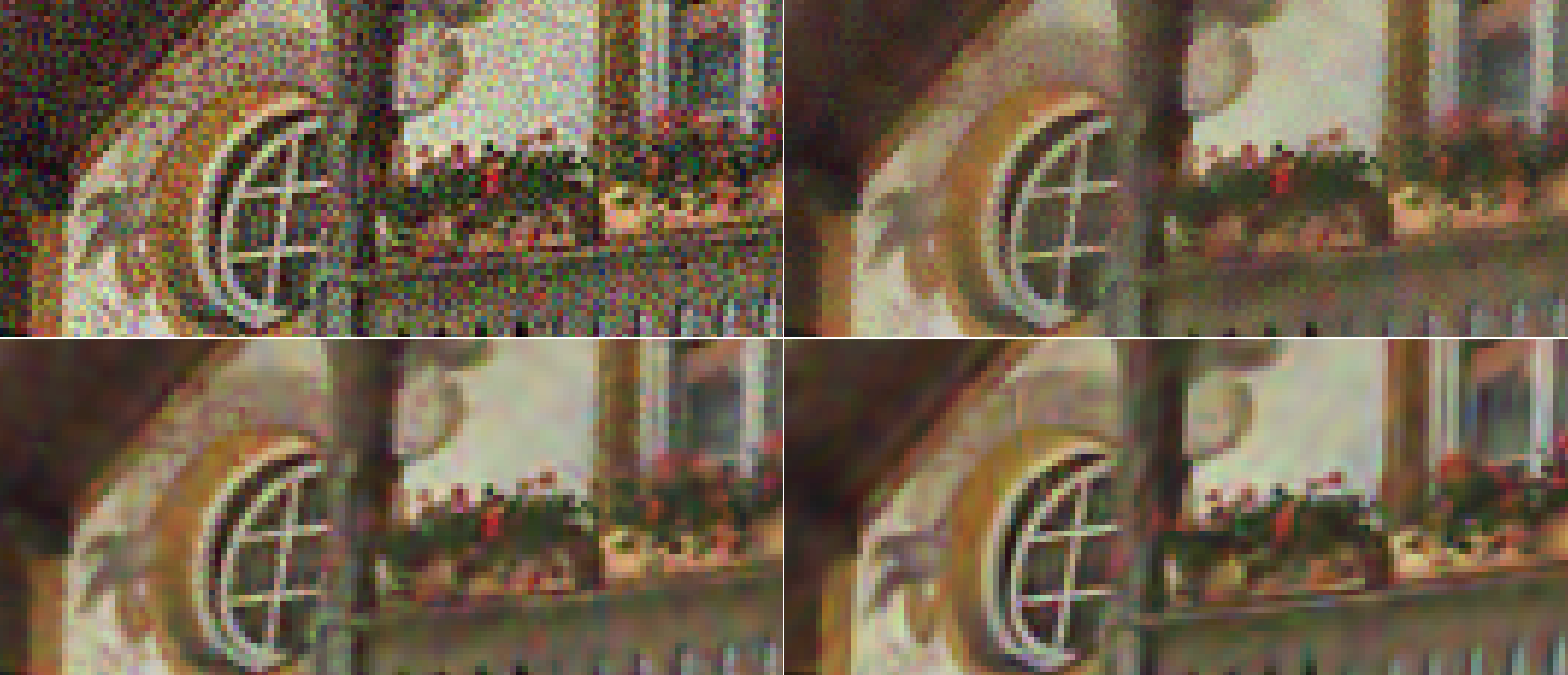}
  \caption{Visual results of anisotropic denoising.
  \textbf{Top left:} noisy input image. \textbf{Top right:} globally optimal 5x5 bilateral filter. \textbf{Bottom left:} PKN predicting isotropic Gaussian sigma. \textbf{Bottom Right:} PKN predicting anisotropic Gaussian covariance.}
  \label{fig:aniso_denoise_qualitative}
\end{figure*}

We parameterize network outputs as two channels with sigma multipliers in the range [0, 4], and a correlation coefficient in the range [-1, 1].
For verification and ablation studies of our experiments, we train also a second network that outputs only a single sigma shared by both directions (equivalent to isotropic blur).
We present qualitative results in \autoref{fig:aniso_denoise_qualitative} and illustrate the inferred coefficients in \autoref{fig:isotropic_anisotropic}.

\begin{Figure}
 \centering
 \vspace{0.1in}
 \includegraphics[width=0.9\linewidth]{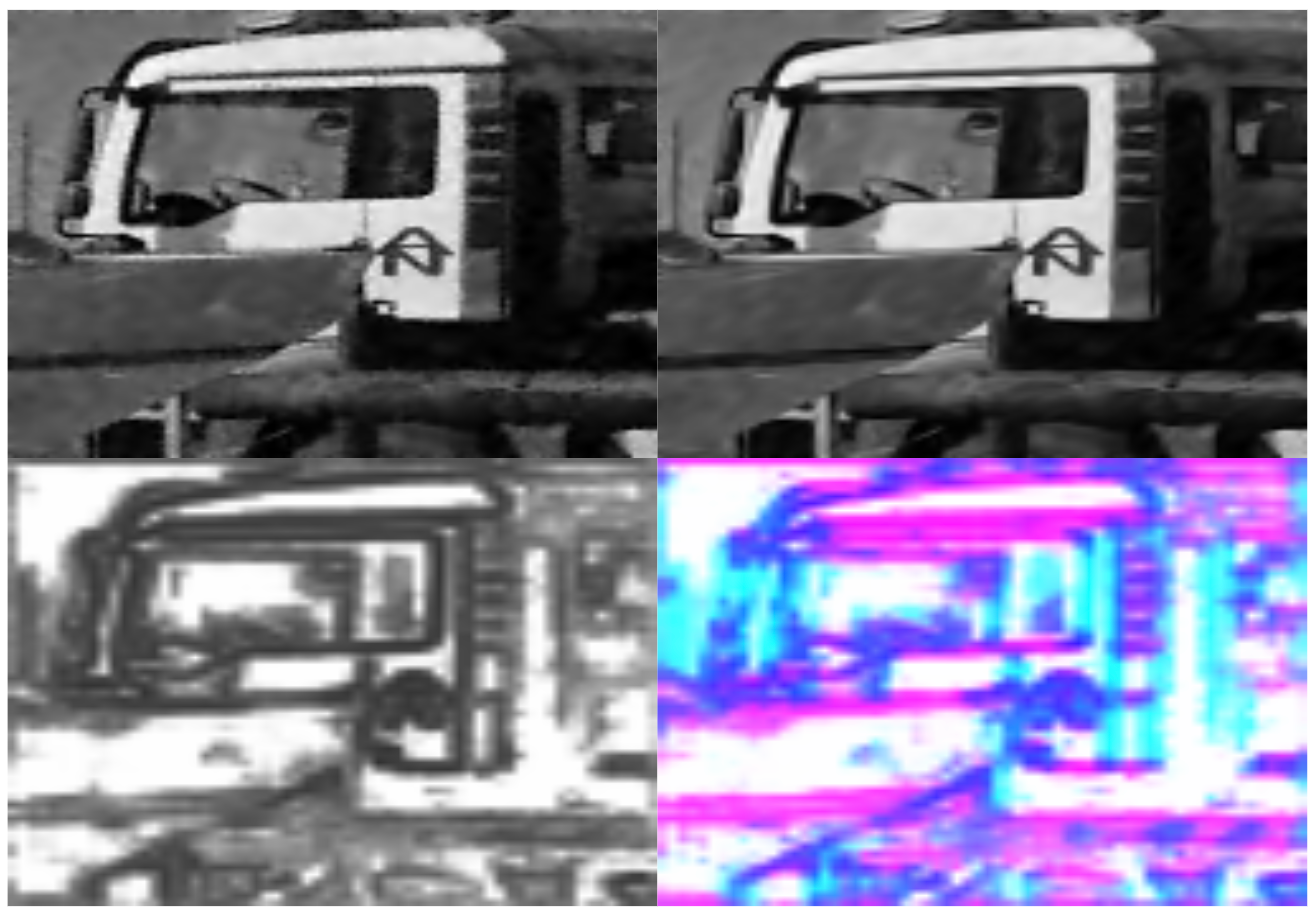}
 \captionof{figure}{
  Denoising with isotropic (\textbf{left}) and anisotropic (\textbf{right}) Gaussian kernels.
  \textbf{Top:} Denoised image result. \textbf{Bottom:} Predicted kernel parameters.
  The anisotropic kernel predicts a precision matrix that visually aligns with edges of different orientations in the image.
  }
 \label{fig:isotropic_anisotropic}
 \vspace{0.1in}
\end{Figure}
Even such a simple network is able to find image features like edges and textures, and denoise the image while preserving salient features.

Anisotropic Gaussian kernels can be used for other applications beyond image processing including 3D and geometry processing, but we don't explore them in this work.
We suggest that for future work, it could be beneficial to analyze the performance of a predicted full anisotropic bilateral kernel characterized by a full 3x3 spatial-signal covariance matrix.
\subsection{Polynomial deblurring kernel}
Prior work suggests to use a prediction of coefficients for a linear combination of bases for tasks like denoising or single frame upsampling \cite{xia2020basis, li2020lapar, carbajal2021single, kokkinos2019pixel}.
We analyze how some of those approaches and applications can be unified under procedural kernel prediction network framework.
Recent work \cite{delbracio2020polyblur} has demonstrated how polynomial reblurring kernel can be used for efficiently approximating deconvolution.

A spatially varying blur kernel can be used to not only prevent amplifying noise while deconvolving the image, but also remove some of it.
Inspired by this work, we use the same polynomial reblurring kernel, but instead of fixed global tuning of polynomial coefficients, we use a network to predict them and adapt locally.

Given blur kernel matrix $B$, a polynomial reblurring kernel is defined as:
\begin{equation}
W(x) = a_x I + b_x B + c_x B^2 + d_x B^3 + ...
\end{equation}
We use a fourth degree polynomial with coefficients $a, b, c, d$.
Those coefficients need to sum to 1 to preserve constant signals, therefore we have only three degrees of freedom and use a network to predict 3 coefficients $b, c, d$, while we parameterize $a$ as:
\begin{equation}
a_x = 1-(b_x + c_x+ d_x)
\end{equation}
For parameters $b, c, d$, we apply a sigmoid remapping function in range [-4, 4] and train for minimization of a L2 loss function.

\begin{figure*}[ht]
  \centering
  \includegraphics[width=0.9\linewidth]{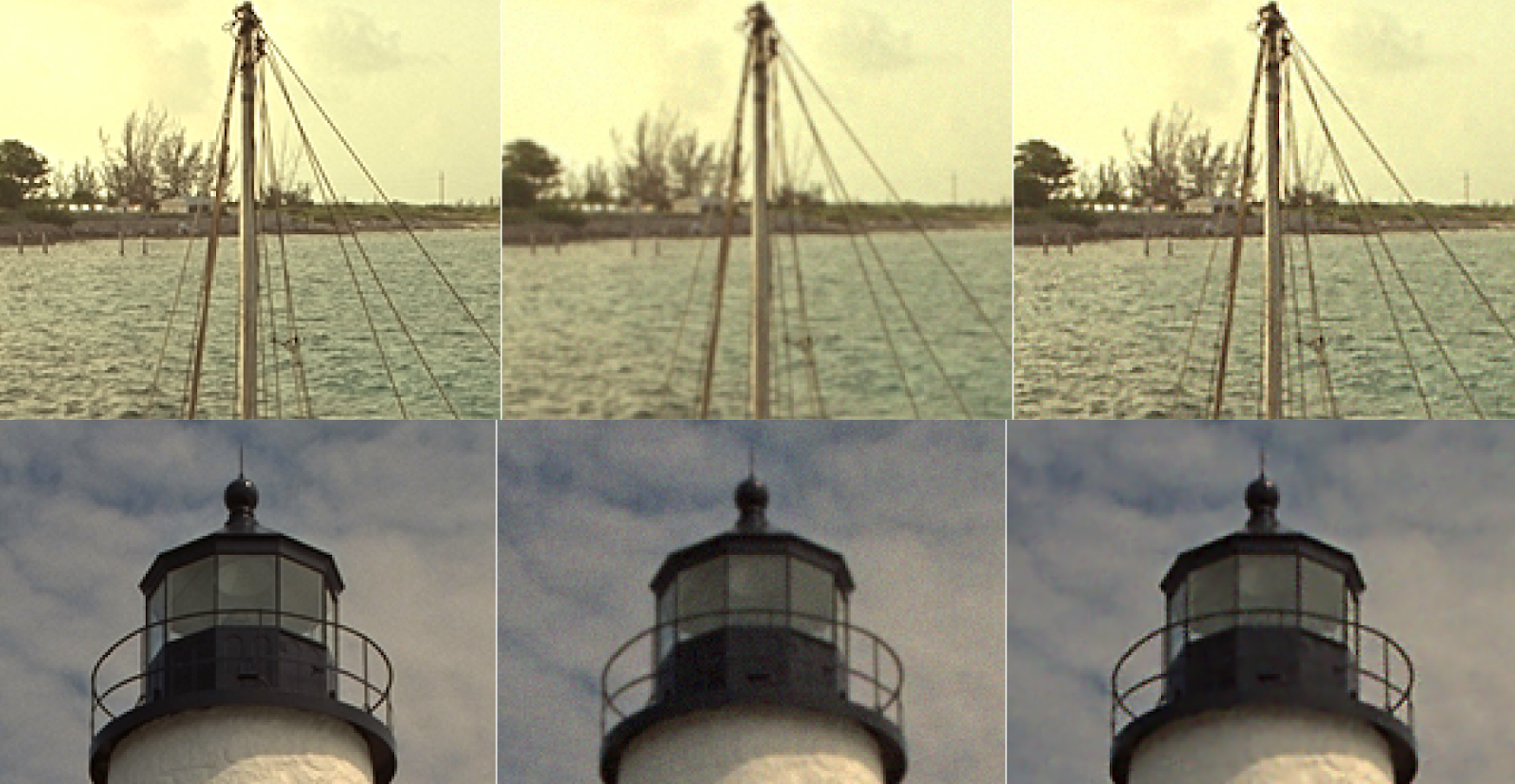}
  \caption{Visual results of polynomial reblurring coefficient prediction.
  \textbf{Left:} Original image crops. \textbf{Middle:} Noisy and blurry images. \textbf{Right:} Results of deconvolution with denoising.}
  \label{fig:reblur_results}
\end{figure*}

As our experiments show (\autoref{fig:reblur_results}), it's possible to achieve both detail amplification and deconvolution of edges, as well as mild denoising effect.

\begin{Figure}
 \centering
 \vspace{0.1in}
 \includegraphics[width=\linewidth]{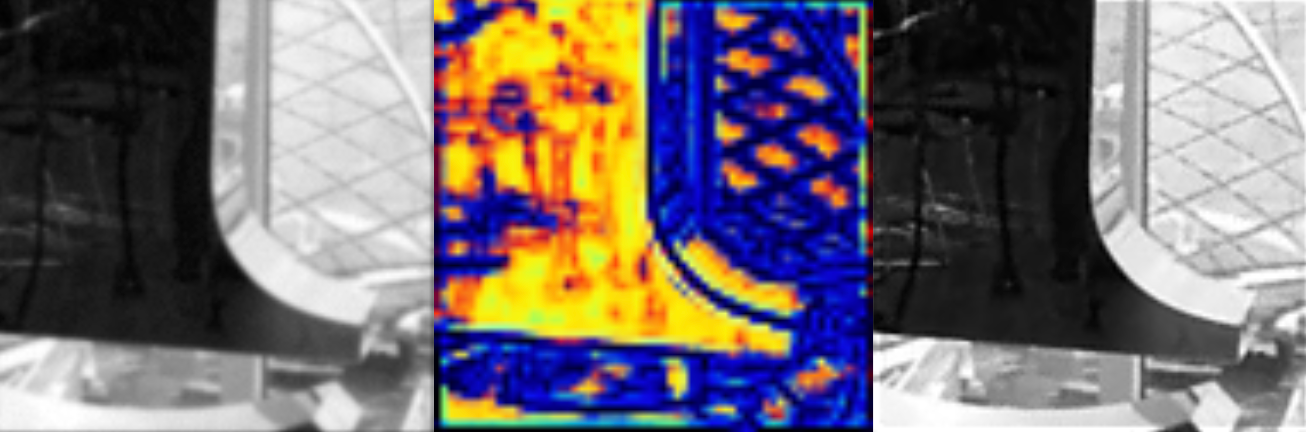}
 \captionof{figure}{
  \textbf{Left:} Original image crop. \textbf{Middle:} Predicted coefficients, where each color channel represents different polynomial coefficient. \textbf{Right:} Deconvolved and denoised image.
  }
 \label{fig:polyblur_coeffs}
 \vspace{0.1in}
\end{Figure}

The resulting predicted coefficients (\autoref{fig:polyblur_coeffs}) are easily interpretable and correspond to detected structures.
In the areas in need of denoising and lacking visible structures, the predicted polynomial forms a simple blur kernel, while in the areas that can benefit from deconvolution, the polynomial performs a kernel inversion.
\subsection{Anisotropic upsampling}
We expand the experiments with anisotropic Gaussian filters in \autoref{aniso_denoise} to single frame upsampling.
Recent work on multi-frame super-resolution \cite{wronski2019handheld} has used anisotropic Gaussian Radial Basis Function kernels for arbitrary, continuous magnification of images.
A continuous parameterization can be used to increase images resolution by different magnification factors, including fractional ones \cite{chen2020learning}.

Similarly to  \cite{wronski2019handheld}, we focus on anisotropic Gaussian RBFs, but instead of using a set of complicated, hand-tuned heuristics applied to a structure tensor computation \cite{bigun1987optimal} described by authors in the supplemental material, we automatically infer them through a neural networks.

We predict anisotropic Gaussian covariance matrix at the half of the original image resolution parameterized the same way as in \autoref{aniso_denoise}.
For simplicity of evaluation, we bilinearly upsample the Gaussian kernel precision matrix to the desired resolution and use as local kernel parameters.

\begin{figure*}[ht]
  \centering
  \includegraphics[width=0.85\linewidth]{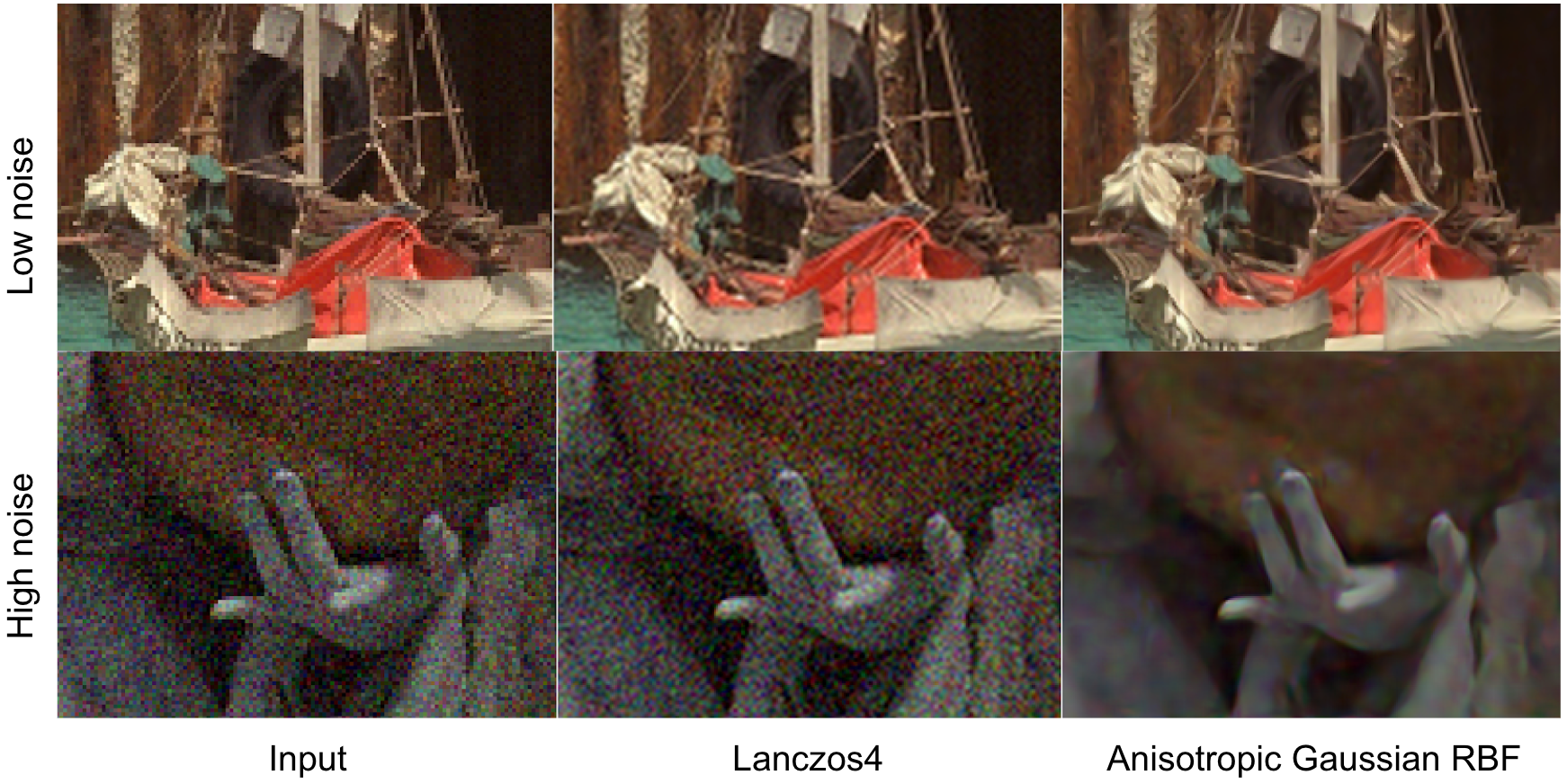}
  \caption{Qualitative evaluation of anisotropic kernel upsampling.
  }
  \label{fig:upsample_results}
\end{figure*}

We train the upsampling network with varying amounts of noise, and compare to a commonly used Lanczos4 upsampling method, showing perceptually better results for both mild and severe noise (\autoref{fig:upsample_results}).

\begin{Figure}
 \centering
 \vspace{0.1in}
 \includegraphics[width=\linewidth]{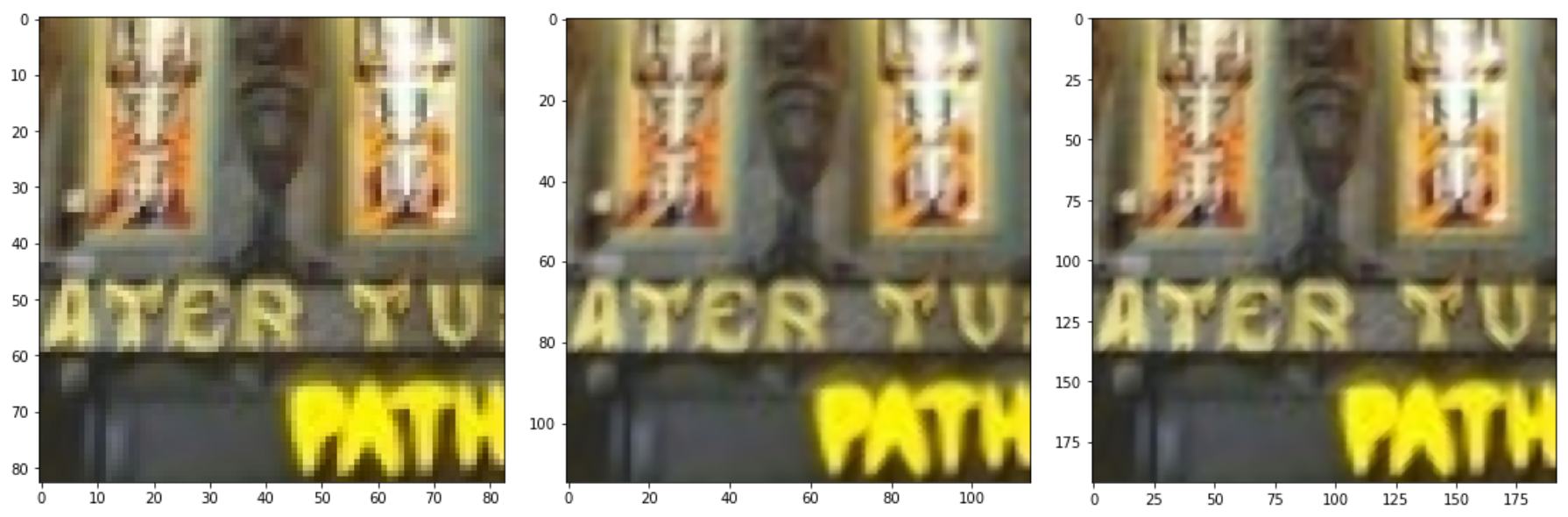}
 \captionof{figure}{
  \textbf{Different upsampling factors:} 1.3x, 1.8x, 3.0x. Readers are encouraged to zoom in in the digital copy.
  }
 \label{fig:upsample_fract}
 \vspace{0.1in}
\end{Figure}

While most of the previous work on learning image upsampling filters is focused on integer, predefined upsampling factors \cite{romano2016raisr}, we can easily upsample by fractional amounts (\autoref{fig:upsample_fract}) to provide an option of fractional zoom to the user.
We note that the inference resolution, or the inference results don't change between the magnification factors, allowing to save or cache the inferred results while the user can preview the image at different magnification factors, potentially even changing in real time.

\section{Discussion and limitations}
In this section we will discuss the results, target uses of Procedural Kernel Networks, as well as some limitations compared to prior work.
\subsection{Kernel limitations}
We observe that our proposed framework is limited by the \textit{oracle} performance of the used function generating kernel for the selected tasks.
While some kernel functions have enough degrees of freedom to predict even the target pixels directly, it's not the case for most of analyzed and proposed uses.

PKN cannot achieve results that are any better than the best possible potential (local) parameters of the given kernel, similar to the oracle bilateral kernel experiments (\autoref{table:psnr_denoise}).

For some uses, this approach produces results that are inferior to the state of the art, but can be still attractive for practical application and on-device use.

\subsection{Quantitative and qualitative performance}
For the reason described above, we don't analyze quantitative performance on most of the proposed tasks.
Even for the denoising application, we note that achieved performance is significantly lower than many state of the art results.

The goal of our work is to show a scalable framework that can be applied to existing building blocks from the vast body of existing research predating the machine learning revolution.
We leverage the capabilities of neural networks to boost the performance of these existing algorithms.
The qualitative results shown confirm that we can produce visually attractive results that are significantly better than the baseline (spatially uniform kernel parameters).
\subsection{Automatic tuning and boosting of existing algorithms}
Our work can be seen as a framework for automatically tuning existing traditional algorithms.
Unlike prior work \cite{tseng2019hyperparameter}, we propose to find those parameters during the inference time of the algorithm, not in an offline optimization process.
The inferred parameter maps are not just globally optimal (best performance averaged over the whole dataset), but also close to locally optimal.

One particularly interesting potential application of our work is that it can be used to drive parameters of {\em existing hardware blocks}, which are typically designed many years in advance and are more difficult to modify and improve than software and algorithm blocks.
Even if the hardware blocks lack the dedicated local parameter inputs, some of the benefits can be achieved through global parameter manipulation, or even manipulating the input---such as by modifying the expected per-pixel standard deviation of the noise.
\subsection{Computation resolution}
We don't assume any particular operating resolution of the kernel generating function.
This is one of the advantages of our formulation, as by using smoothness of the parameter space, one can decouple the operating resolution of the network.
While we don't explore it in our work, if at least a partial correlation of the optimal parameters with the signal strength exists, one could infer at even smaller resolutions, e.g. using a bilateral grid \cite{chen2007real}.

Taken to the extreme, our method can still predict single set of global parameters that are optimal per image, which can be an improvement over finding parameters that are optimal on average for the whole training set.

\subsection{Loss function}
In all of our presented experiments, we have used the L2 loss function.
Depending on the application, one could use a different loss function, for example one that correlates better with the human perception  \cite{zhang2018perceptual}, or other desired properties for applications that are not tied to viewing images by humans, like object recognition or classification.

\subsection{Differentiability and parameterization}
Throughout our work, we conveniently use continuous, differentiable kernels and loss functions.
We have not done any experiments with using differentiable proxies for non-differentiable tasks, such as computing a loss after JPEG compression \cite{shin2017jpeg}, but we believe the proposed approach could apply to such problems.
On the other hand, we observe how some parameterizations of the kernel generating functions are much better suited for learning and the optimization process than the others.

In the case of anisotropic Gaussian kernel prediction, we noted that a direct covariance matrix representation can easily produce an invalid and non-invertible covariance matrices, which leads to exploding gradients and instability during training.
Reparameterizing the matrix using correlation and standard deviation coefficients leads to stable convergence at training time.
We believe that while the loss function can be non-convex in the kernel parameter space, it is required to be relatively smooth and Lipschitz-continuous.
\subsection{Use beyond images}
Finally, we suggest that this work could be expanded beyond the image domain.
Experiments with fractional, continuous upsampling suggest that predicting local functional parameters could work well with irregular or sparse inputs and applications such as meshes, point clouds, or even graphs.
\section{Summary}
To conclude, we have presented a general framework of \textbf{Procedural Kernel Networks} that unifies some of the existing work on real-time and computationally efficient deep learning methods.
We propose how to use this framework with a wide range of different procedural kernel functions.
We propose some practical applications and suggest further work in under-explored areas such as using small neural networks to boost the quality of traditional algorithms or existing hardware blocks.

While our work has limitations and we don't claim to achieve state of the art on any of the analyzed applications, we hope that the vast design space possible in this framework and an opportunity to base on existing research predating the deep learning revolution could inspire some novel applications.
\section{Acknowledgements}
We would like to thank Sam Hasinoff, Cengiz \"{O}ztireli, Manfred Ernst, Dillon Sharlet, Mauricio Delbracio, Julian Iseringhausen, and Jon Barron for valuable discussions and suggestions that lead to this work and the manuscript.
\bibliographystyle{ieeetr}
\bibliography{references}

\end{multicols}
\end{document}